\documentclass[a4paper, 10pt, conference]{ieeeconf}      
\usepackage{FG2025}

\FGfinalcopy 

\IEEEoverridecommandlockouts                              
\usepackage{graphicx}
\usepackage{booktabs}
\usepackage{hyperref}

\def\FGPaperID{253} 

\title{\LARGE \bf
Transfer Learning from Visual Speech Recognition to Mouthing Recognition in German Sign Language
}

\author{\parbox{16cm}{\centering
    {\large Dinh Nam Pham and Eleftherios Avramidis}\\
    {\normalsize
    Speech and Language Technology Department, German Research Center for Artificial Intelligence (DFKI), Berlin, Germany\\}}
}

\usepackage{fancyhdr}
\begin{document}

\ifFGfinal
\thispagestyle{empty}
\pagestyle{empty}
\else
\author{Anonymous FG2025 submission\\ Paper ID \FGPaperID \\}
\pagestyle{plain}
\fi
\maketitle

\thispagestyle{fancy}

\begin{abstract}

Sign Language Recognition (SLR) systems primarily focus on manual gestures, but non-manual features such as mouth movements, specifically mouthing, provide valuable linguistic information. This work directly classifies mouthing instances to their corresponding words in the spoken language while exploring the potential of transfer learning from Visual Speech Recognition (VSR) to mouthing recognition in German Sign Language. We leverage three VSR datasets: one in English, one in German with unrelated words and one in German containing the same target words as the mouthing dataset, to investigate the impact of task similarity in this setting. Our results demonstrate that multi-task learning improves both mouthing recognition and VSR accuracy as well as model robustness, suggesting that mouthing recognition should be treated as a distinct but related task to VSR. This research contributes to the field of SLR by proposing knowledge transfer from VSR to SLR datasets with limited mouthing annotations.

\end{abstract}

\section{INTRODUCTION}

As sign languages (SLs) are fully-fledged visual-manual natural languages, they
are perceived visually and expressed using manual gestures (hand movements) as well
as non-manual signals such as facial expressions, body posture and eye gaze. 
Therefore, SLs serve as an important communication method for the deaf and hard-of-hearing community.
Consequently, automatic sign language recognition (ASLR) has attracted increasing attention from the 
research community to facilitate communication between SL users and non-users \cite{koller2020quantitativesurveystateart}.

While ASLR systems primarily work with manual gestures, utilizing the non-manual features 
has become an emerging trend. These signals can provide valuable information as they are an
essential component in SLs, often compared to intonation in spoken languages \cite{Crashborn2006}.
One of these non-manual markers is mouth movements.
Despite only 5\% of the published SLR results from 2015 to 2020 incorporating the signer mouth characteristics \cite{koller2020quantitativesurveystateart},
their inclusion in ASLR models has been shown to lead to greater performance \cite{pham2023esann, Agris2008, schmidt-etal-2013-using}.
In SL, mouth actions can be categorized into two types: mouth gestures, which are independent of spoken language,
and mouthings, which refers to silently pronouncing a word of the spoken language or at least its first syllable.
Since mouthings are included in almost all studied SLs and play a significant role in both the formal structure and semantic expression of these languages \cite{Bauer2022}, we believe that mouthings should be integrated
in future ASLR systems. In this work, our aim is to contribute to the advancement of mouthing recognition.

One of the key challenges in mouthing recognition is the scarcity of annotated data needed to develop robust models. 
This is largely due to the high cost and time-intensive nature of expert-driven manual annotation, coupled with
the limited number of studies and research efforts dedicated specifically to mouthing.
As mentioned before, mouthings are related to the articulation of spoken words. 
Therefore, as a potential solution to address the scarcity of annotated data for mouthing recognition, we propose leveraging datasets from visual speech recognition (VSR), also known as lipreading, which is more widely studied than automatic mouthing recognition. VSR datasets focus on capturing the articulation of words through mouth movements, which aligns closely with the objectives of mouthing recognition. In this work, we explore transfer learning from VSR to mouthing recognition as a strategy to improve performance and mitigate data limitations. Specifically, we utilize three lipreading datasets: one in English, one in German with words unrelated to the target mouthings, and one in German containing the same target words as the German Sign Language mouthing dataset we created. This setup allows us to investigate how varying levels of relatedness between lipreading datasets and the target mouthing task affect recognition performance. To facilitate this analysis, we employ three different transfer learning approaches: fine-tuning, domain adaptation, and multi-task learning, providing a comparison of their effectiveness in this context.
To the best of our knowledge, this work represents the first attempt to: (1) use the corresponding words of the spoken language as labels for mouthing recognition and (2) apply transfer learning from visual speech recognition to mouthing recognition, proposing a novel approach to improve mouthing recognition and address the challenge posed by limited annotated data.

\section{RELATED WORKS}

A limited number of studies have explored the use of VSR methods to enhance sign language recognition. 
A very brief survey on this topic, done in \cite{Antonakos}, identified two approaches originating from VSR that could 
be applied: (a) recognizing specific words or phrases or (b) recognizing a set of predefined mouth shapes or mouth dynamics
to produce words. One of these two approaches is commonly adopted in most related works.
In \cite{schmidt-etal-2013-using}, a viseme-based mouthing recognizer was incorporated into
a German Sign Language translation framework, outperforming the baseline system that does not utilize mouthing
as an additional knowledge source. Instead of visemes, mouthing annotations describing the mouth shape
were used in \cite{saenz-2022-mouthing}, running an American Sign Language (ASL) dataset through OpenPose \cite{openpose}, a pre-trained CNN-based 2D pose estimator. 
Examples of these mouthing annotations include "open and corners down", "raised upper lip" and "lips spread and corners down". In contrast, we perform mouthing recognition as the task of assigning videos of mouthings to their corresponding spoken words. Furthermore, the frequency of mouthings varies across SLs, with mouthings occurring more often in German Sign Language (DGS) than in ASL \cite{Crashborn2006}. We focus on mouthing recognition in DGS and use German spoken words as labels in the experiments.
A framework for recognizing mouthings in continuous DGS in a weakly supervised manner, utilizing speech transcripts, was proposed in \cite{Kollerreadmylips}. This represents the first use of viseme recognition not only in DGS, but also within the context of sign language recognition. Additionally, \cite{koller:14031:sign-lang:lrec} introduced an automated method for annotating mouthings in DGS, requiring both gloss annotations and speech transcripts. In \cite{Albanie2020}, mouthing was used to facilitate SL subtitle annotation.

In addition to mouthings, research focusing on mouth gestures and mouth actions in DGS exists as well. 
Mouth gestures were classified by training a model on isolated video clips in \cite{brumm:20020:sign-lang:lrec}. To address homonym disambiguation in DGS, \cite{pham2023esann} examined the impact of including mouth actions as an input on model performance.
Moreover, VSR for isolated spoken words in German was done in \cite{pham2022dev} and \cite{schwiebert-etal-2022-multimodal}.

\section{METHOD}

\subsection{DATASETS}

\subsubsection{Mouthing in German Sign Language}
In order to develop and evaluate a model to recognize mouthing, the creation of a dataset was necessary.
For this, the Public DGS Corpus \cite{dgscorpus_3} was identified as a suitable source
due to its extensive collection of SL videos and accompanying annotations. 
This corpus features videos of signers from various regions across Germany and provides detailed transcriptions,
including annotations for signs, translations, mouth gestures and mouthings, along with their corresponding timestamps.
Using this, we determined the number of occurrences of each mouthing in the whole corpus,
selected 15 mouthings with a sufficient number of instances and extracted all
video clips of these according to the timestamps in the transcripts.
In order to keep the dataset balanced, we randomly selected 500 video clips of each
mouthing to be in the dataset and further split it into training, validation and test sets in an 8:1:1 ratio, keeping the class distribution the same. 
In total, the dataset includes 15 classes, split into 3 sets: the training set includes 400 video clips per class,
while the validation and test sets each contain 50 video clips per class.
Before applying the pre-processing steps described later in this work, each video had an original resolution of 640x360 pixels at 50 frames per second, displaying the signer’s entire upper body and face.

\subsubsection{Visual Speech Recognition}

With the goal of investigating transfer learning from VSR to mouthing recognition, we created 3 VSR datasets. First, we used the "Lip Reading in the Wild" (LRW) dataset \cite{chung}, a popular English VSR dataset containing 500 word classes, each with 800-1000 video clips.
We randomly chose 15 words and split them into training, validation, and test sets, ensuring that each split contained the exact same number of video clips per class as done for the mouthing dataset.

Moreover, the dataset ``German Lips" (GLips) \cite{schwiebert-etal-2022-multimodal} consists of 500 German word classes
with 500 instances each. It is already split in a training, validation and test set with the exact same number of instances per class and split as we have before. For our experiment, we selected 15 random word classes that are unrelated to the labels in the mouthing dataset and 15 classes that match the mouthing classes. In other words, these 15 classes correspond to the spoken words associated with the mouthings in the mouthing dataset.
Henceforth, we annotate the created datasets as follows:

\begin{itemize}
  \item $M$ - dataset with mouthings from DGS
  \item $GLips_M$ - German VSR dataset with words corresponding to the mouthings of $M$
  \item $GLips_R$ - German VSR dataset with words unrelated to $M$
  \item $LRW$ - the English VSR dataset
\end{itemize}

\subsubsection{Pre-Processing}

After manual inspection, we discovered that some video clips in the training split of both $GLips_M$ and $GLips_R$ datasets contained no visible face. The class most affected in the training split had 3 such instances, which led us to remove such samples. Additionally, to maintain consistency across all datasets, we randomly removed instances from the training splits of the other datasets. As a result, each of the four datasets now contains 397 instances in the training set per class and 50 instances each in the validation and test set. Thus, the number of instances per class and split is equal in all 4 datasets and all datasets have the same size.

Naturally, the videos differ in the number of frames which is why we standardize all video clips to 30 frames by repeatedly appending the last frame of each video. Furthermore, we crop all videos to the mouth region with a size of 96 x 96 pixels using the implementation of \cite{ma2022visual}. 

\subsection{MODELS}

\begin{figure*}
    \centering
    \includegraphics[width=0.9\textwidth]{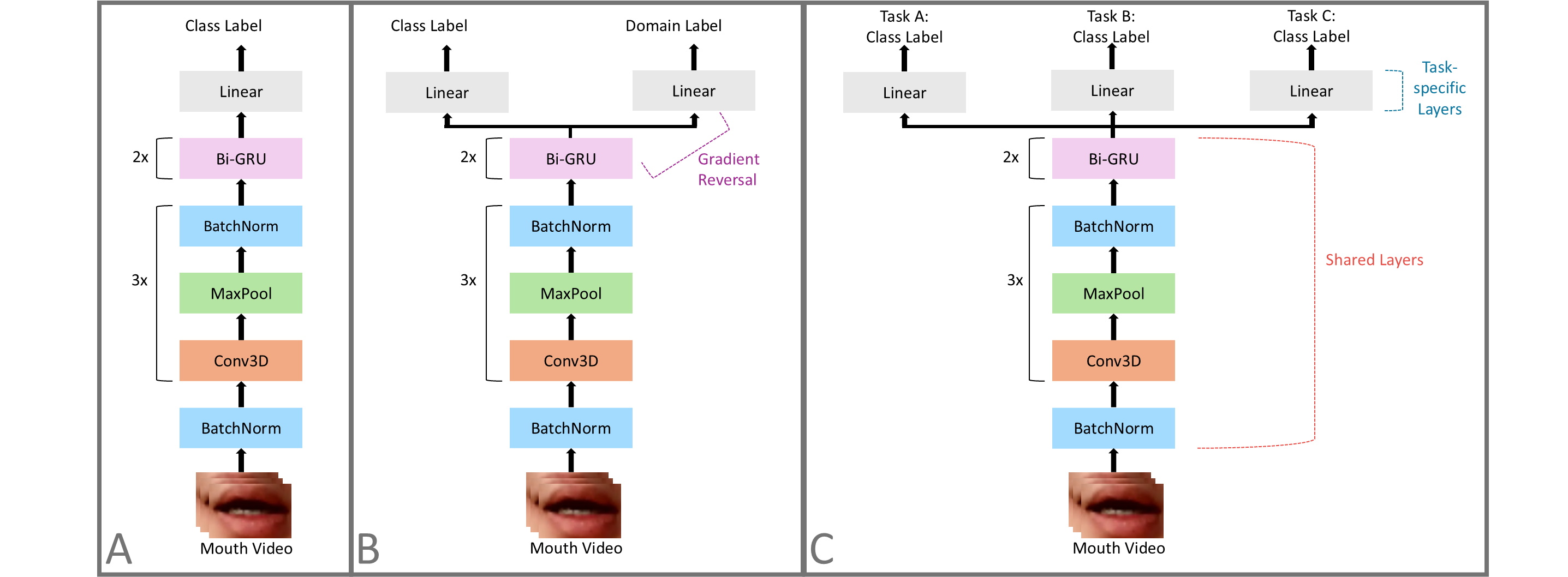}
    \caption{Overview of the model architectures with (A) the baseline, (B) the Domain Adversarial Neural Network and (C) the multi-task learning model.}
    \label{fig:img1}
\end{figure*}

In this section, we explain the architectures of the models we use for the experiments\footnote{The code is publicly available: \url{https://github.com/NPhamDinh/transfer-learning-vsr-mouthing-sign-language}}.
An overview is given in Fig. \ref{fig:img1}.

\subsubsection{Baseline}

To establish a foundation for comparison, we first implement a baseline model for the mouthing recognition task, serving as a reference point for evaluating the transfer learning approaches. The other models will be based on this architecture.
The baseline is an artificial neural network consisting of 3D convolutional layers (Conv3D), bidirectional gated recurrent units (Bi-GRU) and a linear layer for the classification. 
To be more specific, the input video first undergoes batch normalization, followed by three sequential blocks, each consisting of a Conv3D layer, max pooling and batch normalization. These blocks are succeeded by two Bi-GRU layers and a final linear layer. All three Conv3D layers and max pooling operations use a padding of (1,2,2), which also serves as the stride for all max pooling layers. The first two Conv3D layers use 16 channels, while the third employs 32 filters. The first Conv3D layer applies a stride of (1,2,2), whereas the remaining two use a stride of (1,1,1). Each Conv3D layer utilizes a kernel size of (3,5,5), and the Bi-GRU layers have a hidden size of 256.

This architecture is inspired by previous works \cite{assael2016lipnetendtoendsentencelevellipreading, pham2022dev, pham2023esann}, which have demonstrated the effectiveness of similar designs for English and German automatic lipreading, as well as for recognizing mouth actions in DGS. The use of Conv3D layers for feature extraction, combined with recurrent neural networks like GRUs for classification, is well-established and widely adopted in visual speech recognition \cite{Fenghour2021, Sheng2024}.
We selected this architecture for its strong representation of the VSR field, its proven effectiveness and its simplicity. The straightforward implementation and clarity make it a suitable base to build and compare other approaches upon.

\subsubsection{Domain Adversarial Neural Network}

The Domain Adversarial Neural Network (DANN), as proposed in \cite{Ganin2016}, aims to learn a domain-invariant representation through an adversarial training process. To this end, the model consists of two classifiers: one predicts the class and one predicts the domain. During training, the loss for the class prediction is minimized, while the loss for the domain classifier is maximized using a gradient reversal layer. In other words, the model can be seen as learning dataset-independent features that are valuable for predicting the class. In our case, the two domains we will use are $M$ and $GLips_M$, as they share the same labels. Each batch will consist of samples from both domains in equal proportions. The DANN approach is worth exploring to observe the extent to which the shift between the articulation of a word in spoken language and in mouthing can be modeled as a domain shift or whether they should instead be treated as entirely different tasks. This is particularly relevant since many mouthings only articulate parts of the word.

\subsubsection{Multi-Task Learning}

In multi-task learning (MTL), the objective is to learn $n$ tasks jointly to improve the performance of each task by leveraging shared knowledge across tasks. Among the numerous different MTL approaches \cite{Zhang2022MTL}, we chose to implement hard parameter sharing, one of the most widely adopted approaches due to its simplicity and effectiveness. The model has shared layers that learn a common representation across tasks and task-specific layers that are independently optimized for each task. Hence, a common feature representation is being learned that generalizes for all tasks, taking advantage of the relatedness of the tasks. As we will treat each dataset as its own task, we use MTL to explore the task relatedness between $M$ and the different VSR datasets as well as the possible benefits of sharing a feature representation across them. One could argue that for $M$, $GLips_M$ is the most similar task with the same target vocabulary and $LRW$ is the most unrelated task as its words are from an entirely different language. While shared representations can improve generalization across datasets, there is also a risk of performance degradation due to task conflicts. This can occur when the feature representation learned for one task interferes with the optimal representation for another. Through this MTL approach, we aim to explore whether shared representations can effectively capture the relationship between mouthing and visual speech recognition, while experimenting with different sets of tasks with presumably varying degrees of relatedness to DGS mouthing.

                                  
\subsection{EXPERIMENTS}

The baseline architecture will be used to train a model on each of the four datasets individually. Subsequently, the weights of the models trained on the VSR datasets will be fine-tuned on the $M$ dataset without freezing any layers and the final fully connected layer will be re-initialized.

As previously mentioned, DANN will be trained on both $M$ and $GLips_M$, treating both as different domains. For every subset $A \subseteq \{GLips_M, GLips_R, LRW\}$ with $ A \neq \emptyset$, a MTL model will be jointly trained on $A$ and $M$.

All experiments use a batch size of 64, cross-entropy loss as error function and the Adam optimizer with a learning rate of $10^{-5}$. During training, RandAugment \cite{Cubuk2020} is applied to the video input as a data augmentation method. For DANN and MTL models, the total loss is calculated as the sum of all classifier losses, with equal weighting assigned to each. Every model is trained for a maximum of 1500 epochs, with early stopping triggered if the validation accuracy for $M$ does not improve for over 100 epochs after surpassing 1000 epochs.

For evaluation on the test sets, we use the weights from the epoch where a model achieved the highest validation accuracy for $M$. The evaluation is conducted on the test set of $M$ as well as on the VSR test sets, depending on the datasets the model was trained on in the cases of DANN and MTL. To further assess the models' generalization capabilities, we created a perturbed version of the $M$ test set, referred to as $\overline{M}$. It serves to evaluate the models' robustness - how well they perform when faced with unseen data under unexpected or adversarial conditions. The perturbations are generated by applying Gaussian noise and histogram equalization to the $M$ test set. These operations are unseen as they are not part of the RandAugment implementation \cite{fan2021pytorchvideo} used during training.

\section{RESULTS AND DISCUSSION}


\begin{table}[t]
\centering
\caption{Top-1 Accuracies of the models on the test sets.}
\scriptsize
\begin{tabular}{p{3.6cm}p{0.3cm}p{0.3cm}p{0.7cm}p{0.7cm}p{0.3cm}}
\toprule
Model & $M$   & $\overline{M}$ & $GLips_M$      & $GLips_R$      & $LRW$         \\ \midrule
Baseline: $M$                                & 44.00 & 34.67         & -              & -              & -              \\
Baseline: $GLips_M$                           & -  & -            & 38.18          & -              & -              \\
Baseline: $GLips_R$                           & - & -              & -              & 41.47          & -              \\
Baseline: $LRW$                               & -  & -            & -              & -              & \textbf{83.87} \\ \midrule
Baseline: $GLips_M$ $\rightarrow$ $M$     & 45.20 & 40.53          & -              & -              & -              \\
Baseline: $GLips_R$ $\rightarrow$ $M$     & 43.60 & 35.07         & -              & -              & -              \\
Baseline: $LRW$ $\rightarrow$ $M$         & 44.67 & 35.47         & -              & -              & -              \\ \midrule
DANN: $M$ \& $GLips_M$                      & 43.07 & 37.87         & 36.05          & -              & -              \\ \midrule
MTL: $M$ \& $GLips_M$                       & 45.33  & 37.60        & 37.92          & -              & -              \\
MTL: $M$ \& $GLips_R$                       & \textbf{46.53} & \textbf{41.07} & -              & 41.60          & -              \\
MTL: $M$ \& $LRW$                       & 44.80  & 38.93        & -              & -              & 81.60          \\ \midrule
MTL: $M$ \& $GLips_M$ \& $GLips_R$          & 43.33  & 37.73         & 38.85          & 43.20          & -              \\
MTL: $M$ \& $GLips_M$ \& $LRW$              & 45.60 & 36.27         & 40.05          & -              & 80.00          \\
MTL: $M$ \& $GLips_R$ \& $LRW$              & 44.13  & 38.80        & -              & \textbf{45.20} & 80.93          \\ \midrule
MTL: $M$ \& $GLips_M$ \& $GLips_R$ \& $LRW$ & 42.93  & 36.53        & \textbf{40.72} & 43.87          & 81.60  \\ \bottomrule
\end{tabular}
\label{table:rresults}
\end{table}

Table \ref{table:rresults} shows the accuracies of the models on the test sets. It is striking that the models achieve a much higher accuracy (80\% - 83.87\%) on $LRW$, compared to all other datasets, which was also observed when GLips was first introduced \cite{schwiebert-etal-2022-multimodal} and is arguably due to the far better video quality. Although originally from the same source, the accuracy of the baseline for $GLips_R$ is higher than for $GLips_M$. This difference may be due to the fact that its word classes, which were randomly selected, contain more syllables on average compared to the words of $GLips_M$ and are thus easier to visually distinguish. Fine-tuning VSR models on mouthing provides little benefit on the $M$ test set while the performance gains are more significant on $\overline{M}$. Out of all fine-tuning experiments, using $GLips_M$ as the source performed the best. The close relatedness due to the same word classes might be a reason. However, the DANN model does not outperform the baseline for $M$. Yet, as the accuracy is still on a competitive level, the model seems to have learned useful domain-invariant features to some extent. Moreover, it demonstrates improved robustness as it beats the baseline on the perturbed test set. Nevertheless, as it achieves a worse accuracy on $M$ than the baseline, domain adaptation might be ineffective in this case. Although they have the same vocabulary, the discrepancy between spoken articulation and mouthing might introduce differences in visual features, such as the omission of certain phonemes, extent of lip movement and coarticulation effects. This suggests that rather than merely treating them as the same task of different domains, they should be seen as related, but distinct tasks. DANN could still be a viable option to explore if the target domain has considerably fewer labelled samples than the source domain. Having said that, the MTL model for $M$ and $GLips_M$ outperforms the baseline on both mouthing test sets, possibly indicating that treating mouthing and lipreading as entirely different, but related tasks, is the better approach. Furthermore, the MTL model for $M$ and $Glips_R$ achieves the highest accuracy overall on $M$ and $\overline{M}$, improving the performance and robustness for mouthing recognition significantly. In fact, 5 out of the 7 MTL models outperform the mouthing baseline. While mouthing recognition is the main focus, the MTL models incidentally achieve the highest accuracies for the German VSR datasets, meaning that VSR can benefit from mouthing as well. The MTL model jointly learning all 4 datasets at once leads to the lowest accuracy on $M$ out of all models, suggesting that incorporating too many tasks result in task conflicts and degrade performance. Overall, the MTL models yield the highest performance gains for mouthing and German visual speech recognition, demonstrating their effectiveness, as seen in cross-language speech recognition as well \cite{Huang2013}. Our results suggest that task relatedness does not greatly impact the transfer learning benefits in this context. All transfer learning approaches seemingly improve the robustness as they outperform the baseline on the unseen perturbations. The accuracy improvements could become even more significant if the mouthing dataset were smaller or the VSR datasets were larger since performance gains in transfer learning tend to increase when the source dataset is significantly larger than the target dataset, whereas datasets of similar size often yield limited benefit \cite{entezari2023rolepretrainingdatatransfer, Jiang2021SkeletonAM}. Additionally, exploring alternative loss weighting strategies for the MTL architecture, along with different design choices for shared and task-specific layers, could further improve performance.

\section{CONCLUSION}

In this paper, we perform mouthing recognition for DGS and are, to the best of our knowledge, the first to use the corresponding words of the spoken language as labels. To this end, we explore different transfer learning approaches from VSR to mouthing recognition. Fine-tuning of VSR models provides slight improvements, whereas DANN fails to outperform the baseline. Multi-task learning significantly improves both mouthing recognition and German lipreading, demonstrating the benefit of treating mouthing recognition and lipreading as distinct tasks. Future work should explore alternative domain adaptation, MTL, and hybrid methods to further improve mouthing and sign language recognition.



\section*{ETHICAL IMPACT STATEMENT}
Given the nature of this research, which involves the analysis of publicly available datasets and does not involve human participants, animals, hazardous biological agents or sensitive data, an ethical review was not required. Nonetheless, the research was conducted with careful attention to ethical principles, including data integrity, transparency, and respect for privacy. All data used in this study were obtained from sources that permit their use for research purposes. Although facial information is visible in the videos from the datasets, we do not use any identity-specific data or draw conclusions based on religion, race, or gender. Instead, our analysis focuses on patterns and features without attributing findings to any particular group or individual.

However, we acknowledge the existence of ethical concerns and the possibility of misusing the findings of this work. Bias exists in the datasets, as the vast majority of the speakers and signers are visibly white adults, potentially leaving out people of other demographics in this research. Further research efforts should focus on creating sign language and visual speech recognition datasets that are more diverse, inclusive, and representative of global society, rather than reflecting a predominantly Eurocentric perspective. Moreover, it should be noted that we focused on mouthing in German Sign Language. As such, one should be careful to not misunderstand the generalizability of this work and overgeneralize the results for other sign languages. After all, sign language is not universal and all sign languages can vary greatly, often being mutually unintelligible. Therefore, these differences should be respected and it is important to avoid categorizing all sign language users in one homogeneous group, overlooking their unique characteristics and perspectives.
Additionally, while this research aims to contribute to accessibility, there is a risk that sign language recognition and automatic lip reading models could be misused for surveillance or unauthorized monitoring of individuals. 
To mitigate potential risks, our approach focuses solely on linguistic patterns rather than individual identification, and our models do not infer personal attributes. As mentioned before, we use publicly available datasets without identity-specific annotations. This paper aims to contribute to more accessible and inclusive sign language recognition systems, hoping to bridge communication barriers for the Deaf and hard-of-hearing communities. Hence, we are convinced that the benefits strongly outweigh the potential risks.


{\small
\bibliographystyle{ieee}
\bibliography{egbib}
}

\end{document}